\documentclass{article}
\usepackage{iclr2026_conference,times}
\usepackage{amsmath,amssymb}
\usepackage{booktabs}
\usepackage{graphicx}
\usepackage{multirow}
\usepackage{xcolor}
\usepackage[hidelinks]{hyperref}
\usepackage{url}

\newcommand{\armA}{\textsc{scored}}
\newcommand{\armB}{\textsc{generated}}
\newcommand{\armC}{\textsc{verbal-p}}

\title{Scored vs.\ Generated Readouts in Behavioral Language Models:\\
An Empirical Study of Elicitation Format}

\author{\textbf{Touchapon Kraisingkorn, Krittin Pachtrachai, Wachiravit Modecrua}\\
\normalfont Amity Research and Application Center (ARAC), Amity Group\\
\texttt{\{touchapon,krittin,wachiravit\}@amity.co}}

\iclrfinalcopy
\begin{document}
\maketitle
\lhead{}

\begin{abstract}
Language models fine-tuned on real customer behavior are increasingly deployed as
\emph{behavioral simulators}: given a shopper's history and an offer, they predict whether the
person will act, and, unlike a conventional classifier, can explain the prediction in the
customer's voice. Teams routinely treat these two capabilities as interchangeable, assuming
that letting the model reason before it answers is, at worst, neutral for the prediction
itself --- an assumption with real stakes, since a misranked targeting system silently
misallocates incentive budget, with no sign of the failure in the rationale's fluency. We test
this directly, holding checkpoint and prompt content fixed and varying only whether the answer
is read as a probability (\emph{scored}) or produced after a written rationale
(\emph{generated}), across 13 model$\times$domain cells spanning four retail prediction tasks
in three markets (two using fully public data and public checkpoints). The interchangeability
assumption fails: the scored readout ranks outcomes more accurately than the generated one in
12 of 13 cells (two-sided sign test, $p\approx.003$), by $1.5$ to $14.5$ AUC points, with paired
bootstrap confidence intervals excluding zero in every newly measured cell.
The gap is systematic rather than incidental: it scales with the degree of task-specific
supervision and with mismatch between the training and serving formats, ranging from $-2.2$
(untuned base) to $+13.7$ (rationale-format supervision), with a frontier reasoning model at
$+1.4$ (CI$_{95}$ $[0.6,2.2]$). Per-row analysis of ${\sim}9{,}000$ rationales locates two
correlates: rationales that stop citing the dominant predictive feature (rank correlation with
it falls $0.72\!\to\!0.31$) and that converge on a small set of stock formulations (up to $33\%$
of generations share a top-3 opening). Probability saturation, the most visible symptom, does
not track the gap. A third readout --- eliciting a probability \emph{before} any verdict ---
recovers calibration (Brier $0.47\!\to\!0.15$) while ranking within noise of the scored
readout, but only for outcome rates present in training, and it is \emph{worse} than scoring
for models whose scored head is already calibrated. We interpret the gap through the objective
each readout matches, report which training choices narrow it, and give a serving recipe that
retains generated rationales while sourcing ranking from the scored head.
\end{abstract}

\section{Introduction}
\label{sec:intro}

Retailers and marketplaces increasingly decide who receives a coupon, which customers to
prioritize after a product delisting, or how to size an incentive budget using a single
artifact: a language model fine-tuned on real transaction histories to act as a \emph{behavioral
simulator} that can be asked, in natural language, whether a given customer will act
\citep{modecrua2026lbm,wang2025customerr1,maier2025ssr}. Unlike a conventional classifier, such a
model can also be asked to explain itself --- to write, in the customer's voice, why it expects
that outcome. This dual capability is precisely why behavioral simulators are attractive to
build: one checkpoint appears to deliver both a ranking signal for targeting and an auditable
rationale for the analyst who has to justify the targeting decision.

\textbf{Behavioral simulators as a research trend.} Building a single model that both predicts
and explains is not unique to retail: it follows a broader shift toward using language models as
general-purpose world models and agents, continually pretrained and fine-tuned to internalize
the dynamics of a domain rather than only its labels \citep{qwenagentworld2026}. Applied to
consumer behavior, the recipe is to continue pretraining a base model on a corpus of consumer
voice and transaction sequences and then supervise it on the decision task itself, so that the
same weights that generate a plausible customer narrative can also be queried for a prediction
\citep{modecrua2026lbm,wang2025customerr1}. The appeal over a conventional discriminative model
is threefold: a natural-language interface that product teams and analysts can query directly,
transfer to question forms not seen during training, and --- unlike most agent benchmarks, which
score plausibility or human preference --- a domain in which the model's claims can be checked
against a real, recorded outcome (a redemption, a churn event, a repurchase) rather than a proxy
label. That last property also exposes the paradigm's limits: benchmarks that do compare
simulated and real populations continue to find systematic gaps between them
\citep{simbench2025}. It is against this backdrop --- models that are fluent but not yet known to
be faithful --- that how one queries the model for a decision becomes consequential.

Getting the ranking wrong has a direct, material cost. A targeting system that misorders
customers spends incentive budget on people who would have acted anyway and withholds it from
people who would have responded, and it does so silently: nothing in a rationale's fluency
signals that the ranking beneath it has degraded. Whether a team can trust the
rationale-producing surface for ranking, or must fall back to a separate scoring pass, is
therefore a first-order operational question rather than a modeling nicety. In practice, teams
routinely treat the two as interchangeable, assuming that asking the model to articulate its
reasoning is, at worst, neutral for the prediction it then makes. This assumption is not
unreasonable on its face: rationales generated during training have been shown to improve a
model's own accuracy when used as a bootstrapping signal \citep{star2022}, and verbalized
reasoning can rival or even exceed logprob-based confidence for instruction-tuned models on
other tasks \citep{tian2023justask}, so it is a short step to expect that reasoning before
answering would be neutral, or even helpful, at serving time as well. Yet the same reasoning
step has separately been shown to induce overconfidence as its budget grows
\citep{calibrationdrift2026} and non-monotone accuracy as it lengthens \citep{wu2025moreisless}
--- reason enough to distrust the assumption rather than take it on faith, a tension the
literature below makes precise.

The evidence for the interchangeability assumption is indirect, and it comes from several
adjacent literatures rather than a single line of work that has tested it directly.

\textbf{Evidence on elicitation and reasoning.} How a model is asked is already known to matter
along two related dimensions. First, at the level of a single token: \citet{wang2024myanswerc}
show that a model's first-token probability and its generated free-text answer can disagree on
multiple-choice benchmarks, framing the discrepancy as a threat to evaluation validity rather
than to deployment, and this sensitivity to elicitation protocol recurs broadly in confidence
estimation more generally \citep{protocolsensitivity2026}. Second, at the level of reasoning
itself: chain-of-thought has been shown to induce overconfidence as the reasoning budget grows
\citep{calibrationdrift2026} and to produce non-monotone accuracy as rationales lengthen
\citep{wu2025moreisless}, and \citet{opera2025} found that removing rationales changes
action-prediction quality for some models without isolating why. Taken together, this literature
establishes that elicitation format is not innocuous in general, but none of it holds a
checkpoint and its prompt content fixed while varying only the readout, and none measures the
resulting cost against real, verified behavioral outcomes.

\textbf{Evidence on behavioral simulators specifically.} \citet{modecrua2026lbm} and
\citet{wang2025customerr1} establish the training recipes for behavioral simulators of exactly
the kind we study and report in-domain accuracy under a single, fixed readout, without varying
elicitation format. \citet{tian2023justask} and \citet{maier2025ssr} study
verbalized-probability elicitation --- respectively on general instruction-following tasks and
on survey-style purchase-intent items --- rather than on real, verifiable behavioral outcomes.
That is the gap this paper fills: a controlled, same-checkpoint comparison of elicitation
formats, measured against ground-truth customer behavior rather than a benchmark label or a
synthetic proxy.

This paper tests the interchangeability assumption directly, holding checkpoint and prompt
content fixed and varying only how the model is asked, and finds that it does not hold for tuned
behavioral models. Across four domains, reading a decision probability from the model
(\emph{scored}) ranks real outcomes more accurately than letting the model write a rationale and
then answer (\emph{generated}) --- and the difference is largest in exactly the models trained
hardest on the task.

We stress at the outset what this is and is not. It is a measurement of \emph{ranking accuracy
under a fixed checkpoint}: which readout best recovers the ordering of real outcomes. It is not
a claim that generated rationales lack value. They serve purposes ranking metrics do not
capture --- auditability, interpretability, transfer to unseen question forms, and a product
surface customers and analysts can interrogate --- and in our measurements they are sometimes
\emph{better calibrated} than the scored readout even while ranking worse. Rationale-producing
behavioral models are an active and, in our view, worthwhile research direction; the
contribution here is to quantify one specific cost so that it can be designed around rather
than paid silently.

\textbf{Why a gap should be expected.} The scored readout is the model's native binary-outcome
head: a probability over $\{$\texttt{YES},\texttt{NO}\} conditioned on all prompt evidence,
which is precisely the quantity a discriminative model trained with binary cross-entropy on the
same label optimizes, and precisely what a ranking metric such as AUC rewards. The generated
readout routes the same evidence through a text bottleneck first, then reads a probability
conditioned on that text. Any evidence the prose fails to carry --- or distorts --- cannot be
recovered downstream. Established BCE-trained baselines on these tasks (gradient-boosted trees;
the DMBGN voucher model, \citealp{dmbgn2021}) never route evidence through text at all, and, as
we report, they remain strong. The question this paper answers empirically is how large the
text-bottleneck cost is, what modulates it, and what can be done about it.

This paper makes four contributions:
\begin{enumerate}
\item \textbf{The measurement} (Section~\ref{sec:gap}): across four domains --- coupon
redemption at a major Southeast-Asian grocery retailer, e-commerce voucher redemption on the
public DMBGN benchmark \citep{dmbgn2021}, assortment-change response from natural experiments,
and coupon-campaign redemption on the public Dunnhumby \emph{Complete Journey} dataset
\citep{dunnhumby} --- and five checkpoints ranging from an untuned 27B base to a frontier
reasoning API, the scored readout beats the generated one in 12 of 13 cells; in the most
extreme case, a generated arm that is degenerate (single-class output) is recovered to $0.766$
AUC by changing the readout alone.
\item \textbf{What modulates the gap} (Section~\ref{sec:gap}): it grows with task-specific
supervision and with train/serve format mismatch, from $+9.3$ to $+13.7$ points under
rationale-format supervision, to $+3.0$ points --- the smallest among tuned models --- under
in-domain decision-format supervision, with an untuned base showing the reverse sign ($-2.2$).
\item \textbf{Two correlates and one non-correlate} (Section~\ref{sec:mechanism}): per-row
analysis of the generated rationales shows that anchor abandonment and template convergence
order the gaps correctly across cells, while probability saturation, the most visible symptom,
does not.
\item \textbf{A third readout and its scope} (Section~\ref{sec:repair}): eliciting a probability
before any verdict recovers calibration with ranking within noise of the scored readout, but
only for outcome rates present in training, and it is strictly worse than scoring for models
whose scored head is already calibrated.
\end{enumerate}
Together these findings motivate concrete training and serving guidance (Section~\ref{sec:protocol})
that retains generated rationales in the product while sourcing ranking from the scored head,
together with two diagnostics that predict how much a given deployment is paying for its
readout choice.

The remainder of the paper is organized as follows. Section~\ref{sec:related} situates this
measurement against related work on evaluation-time answer divergence, reasoning calibration,
and verbalized probabilities. Section~\ref{sec:setup} describes the checkpoints, training
pipeline, evaluation domains, and the three readouts we compare. Section~\ref{sec:results}
reports the main measurement (Section~\ref{sec:gap}) and its per-row correlates
(Section~\ref{sec:mechanism}). Section~\ref{sec:discussion} introduces a third readout and
characterizes when it helps (Section~\ref{sec:repair}), and discusses limitations
(Section~\ref{sec:limits}). Section~\ref{sec:conclusion} synthesizes the findings, and
Section~\ref{sec:protocol} closes with deployment guidance. Two of our four domains use public
data and public checkpoints end-to-end, so those numbers are reproducible by third parties;
code, prompts, and per-row dumps are released.

\section{Related work}
\label{sec:related}

\textbf{Readout choice as a measurement problem.} \citet{wang2024myanswerc} showed that
first-token probabilities and generated text answers disagree on multiple-choice benchmarks,
framing it as a threat to \emph{evaluation validity}. We study the deployment counterpart ---
which readout best recovers real-outcome ranking --- and add the supervision dose--response and
per-row correlates that multiple-choice settings cannot expose.

\textbf{Reasoning and calibration.} Chain-of-thought has been shown to induce overconfidence
under increased budgets \citep{calibrationdrift2026} and non-monotone accuracy in length
\citep{wu2025moreisless}. Our observation is distinct and partly opposite in sign: several of
our generated arms are \emph{better calibrated} than their scored counterparts while ranking
worse, so the effect we measure is on ordering, not (only) on scale.
\citet{opera2025} reported that removing rationales improves action prediction for some models
without isolating why; we supply candidate correlates and a controlled ordering ablation.

\textbf{Verbalized probabilities.} Verbalized confidence can rival or exceed logprob-based
confidence for instruction-tuned models \citep{tian2023justask}, and the result is highly
sensitive to elicitation protocol \citep{protocolsensitivity2026}. \citet{maier2025ssr} recover
realistic Likert \emph{distributions} through semantic-similarity elicitation. We extend this
line to real behavioral outcomes and identify two scope conditions: verbalized numbers transfer
to rates seen in training but not to novel aggregate quantities, and they can be strictly worse
than scoring when the scored head is already calibrated.

\textbf{Behavioral simulators.} Customer simulators are built with continued-pretraining,
supervised fine-tuning and RL pipelines \citep{modecrua2026lbm,wang2025customerr1} adapted from
language world-model recipes \citep{qwenagentworld2026}, and benchmarks continue to find gaps
between simulated and real populations \citep{simbench2025}. RL post-training is known to
reduce output diversity \citep{kirk2023rlhf,verbalizedsampling2025}, a plausible contributor to
the template convergence we observe. Our contribution to this line is operational: given any
such model, the serving readout is a first-order determinant of delivered ranking quality, and
supervision \emph{format} determines robustness to readout choice.

\section{Models, training, and data}
\label{sec:setup}

Because the gap we measure depends on how a checkpoint was trained, we describe the training
pipeline in full. All models below are named for \emph{how they were trained}, not by internal
release numbers.

\subsection{Base model and pipeline}
All open-weights checkpoints derive from a single public base, \textbf{Qwen3.5-27B}, through a
three-stage pipeline adapted from language world-model recipes \citep{qwenagentworld2026}:
continued pretraining (CPT) to inject domain dynamics, supervised fine-tuning (SFT) to activate
a response format, and --- for some lines --- preference or RL post-training, which is not
varied in this study. The pipeline and the SEA retail cohort follow the promptable
retail-customer model of \citet{modecrua2026lbm}, scaled here from the 8--9B models reported
there to 27B; that work established the training recipe and evaluated in-domain accuracy, and
does not study elicitation format, which is the subject of this paper. We treat those
checkpoints as given and vary only the readout. All stages use LoRA adapters with under $1\%$ of parameters trainable,
trained with 8-way data-parallel on H200-class hardware; adapters are merged for serving.

\textbf{CPT corpus (SEA line).} Roughly $470$M tokens combining (i) a consumer-voice corpus of
${\sim}1.15$M cleaned documents (${\sim}116$M tokens) --- product reviews, forum and social
discussion, and app-store reviews in Thai, Bahasa and English, deduplicated and
register-preserving; (ii) serialized customer trajectories from a major SEA grocery retailer's
loyalty program: $1{,}500$ customers, $159{,}667$ baskets and $641{,}310$ item-lines over
${\sim}15$ months, rendered as (persona $\to$ occasion $\to$ basket) sequences; and (iii) a
replay mixture capped at $45\%$ to limit forgetting. Ablations during development found that
injecting retrieved product context into the CPT corpus \emph{hurt} downstream accuracy, so CPT
is trained on behavior and voice only.

\textbf{CPT corpus (Western line).} The same construction on Western sources
(${\sim}20.3$M tokens of English consumer voice: forum discussion and app reviews), used for
the checkpoint evaluated on the US grocery domain.

\subsection{The five checkpoints}
We evaluate five checkpoints together with an external reference model. \textbf{Base} is
Qwen3.5-27B with no behavioral training, serving as the untuned reference. \textbf{MultiTask-SFT}
continues from the SEA CPT checkpoint through a $91.7$k-example multi-task SFT mixture in
\emph{decision format} (answer first, no rationale block), covering coupon-acceptance decisions
from real tender-labeled events, churn and repurchase-cycle questions, willingness-to-pay
ladders, and survey-response families. \textbf{Persona-SFT} fine-tunes the same CPT base
instead on a regionally-focused persona corpus (${\sim}19.6$k synthesized personas answering
behavioral and survey items), testing whether persona breadth can substitute for decision
supervision. \textbf{Rationale-SFT} uses the same CPT base and the same $91.7$k mixture,
augmented with $104$k examples regenerated in \emph{rationale format}: a first-person
\texttt{<think>} block precedes the decision token, matching the base model's native thinking
layout, making this the checkpoint whose training most closely matches the generated readout.
\textbf{InDomain-SFT} instead starts from the Western CPT checkpoint and applies decision-format
SFT on the \emph{training split of the target domain itself} ($4{,}890$ US grocery campaign
exposures at natural class marginals, across three question registers), plus a persona lane and
a distribution-calibrated survey lane, isolating what format-matched, in-domain supervision
buys. We additionally evaluate a \textbf{frontier reasoning API model} (\texttt{gpt-5.5}) as an
external reference. Checkpoints for the two public domains are publicly released.

\subsection{Evaluation domains}
All four tasks are binary prediction against \emph{real} outcome labels, with strict temporal
or customer-disjoint splits and leak audits. The \textbf{SEA grocery coupon} domain
($n{=}2{,}000$; base rate $.199$) consists of coupon-tender events inferred from campaign
exposure at a major SEA grocery retailer; training and evaluation customers are disjoint, and
all features are computed strictly before the evaluated event. Its data is proprietary, though
the protocol and code are released. The \textbf{SEA e-commerce voucher} domain ($n{=}3{,}120$,
an evaluation shard of a $12{,}480$-row test set) is the public DMBGN voucher-redemption
benchmark \citep{dmbgn2021}. The \textbf{assortment change} domain ($n{=}380$) captures
customer response to product delistings, mined as natural experiments from raw transactions; we
use only the removal family, since the introduction family of this instrument was withdrawn
after an audit found a feature leak. Finally, the \textbf{US grocery campaign} domain
($n{=}2{,}318$; base rate $.140$) consists of (household, campaign) coupon-redemption exposures
from the public Dunnhumby \emph{Complete Journey} dataset \citep{dunnhumby}, time-split so that
test campaigns
start after all training campaigns, with campaign track-record features computed from
training-period campaigns only; it is rebuilt from a public mirror that reproduces our
instrument's row counts and labels exactly.

\subsection{The three readouts}
All arms share the system prompt (persona, purchase history, campaign facts) and the factual
content of the user prompt; only the elicitation differs. Under \armA{}, the model is
instructed to ``answer exactly YES or NO,'' and the score is $P(\texttt{YES})$ renormalized
over the $\{$\texttt{YES},\texttt{NO}$\}$ token mass at the first generated token, at
temperature $0$. Under \armB{}, the model is instead instructed to reason in first person
inside \texttt{<think>} tags (2--4 sentences); generation is stopped at \texttt{</think>}, the
identical \armA{} cue is then appended, and the score is computed identically --- i.e.\
$P(\texttt{YES}\mid\text{its own rationale})$ --- which isolates the rationale as the only
difference between the two arms. Under \armC{}, the model is asked to ``give your likelihood as
a number 0--100'' \emph{before} any verdict; the score is NN$/100$, read with a lenient parser
(parse-failure rates $\le 0.2\%$ except where noted). For the API model, which exposes neither
token logprobs nor temperature control, \armB{} and \armC{} are realized as prose-then-number
and number-first respectively; \armA{} is not available for this model (Section~\ref{sec:gap}).

\subsection{Baselines and statistics}
Where a gradient-boosted model can be fitted on the same tabular features we report it, and it
wins: leaves-AUC $.838$ versus the best language-model arm's $.766$ on assortment change; a
profile$+$session GBDT at $.866$ in the full-feature voucher setting. This paper is about which
readout best recovers what a behavioral language model encodes, not about superiority over
tabular methods. All newly measured contrasts are paired on identical rows with $10^4$-resample
bootstrap confidence intervals; historical cells cite their original pinned protocols (decoding
mode, shard and seed counts).

\section{Experiments and Results}
\label{sec:results}

We first report the headline measurement across all 13 cells (Section~\ref{sec:gap}), then turn
to what, at the level of individual rationales, explains it (Section~\ref{sec:mechanism}).

\subsection{Scored versus generated}
\label{sec:gap}

\begin{table}[t]
\centering
\caption{Ranking accuracy (AUC) by readout ($n{=}2{,}000$ / $3{,}120$ / $2{,}318$ per domain,
top to bottom). Same checkpoint, same prompt content; only the elicitation differs. $\Delta = $ \armA{} $-$ \armB{} in points; positive means the scored
readout ranks better. Brackets: paired bootstrap CI$_{95}$ where newly measured. Format:
N $=$ no task tuning, D $=$ decision-format SFT, R $=$ rationale-format SFT.}
\label{tab:gap}
\small\setlength{\tabcolsep}{4.5pt}
\begin{tabular}{llccc}
\toprule
Domain & Checkpoint (format) & \armA{} & \armB{} & $\Delta$ (pts) \\
\midrule
\multirow{4}{*}{SEA grocery coupon}
 & Base (N) & .591 & .613 & $-2.2$ \\
 & MultiTask-SFT (D) & .697 & .682 & $+1.5$ \\
 & Persona-SFT (D) & .626 & .531 & $+9.5$ \\
 & Rationale-SFT (R) & \textbf{.721} & .584 & $\mathbf{+13.7}$ \\
\midrule
\multirow{4}{*}{SEA e-comm.\ voucher}
 & Base (N) & .637 & .612 & $+2.5$ \\
 & MultiTask-SFT (D) & .664 & .580 & $+8.4$ \\
 & Persona-SFT (D) & .664 & .519 & $+14.5$ \\
 & Rationale-SFT (R) & \textbf{.683} & .558 & $+12.5$ \\
\midrule
\multirow{3}{*}{US grocery campaign}
 & Base (N) & .786 & .747 & $+3.9$ {\scriptsize$[1.5,6.4]$} \\
 & InDomain-SFT (D) & \textbf{.808} & .778 & $+3.0$ {\scriptsize$[0.9,5.2]$} \\
 & Rationale-SFT (R) & .790 & .698 & $\mathbf{+9.3}$ {\scriptsize$[6.3,12.2]$} \\
\midrule
US grocery (API)
 & \texttt{gpt-5.5} number- vs.\ prose-first & \textbf{.824} & .810 & $+1.4$ {\scriptsize$[0.6,2.2]$} \\
\bottomrule
\end{tabular}
\end{table}

\begin{figure}[t]
\centering
\includegraphics[width=\linewidth]{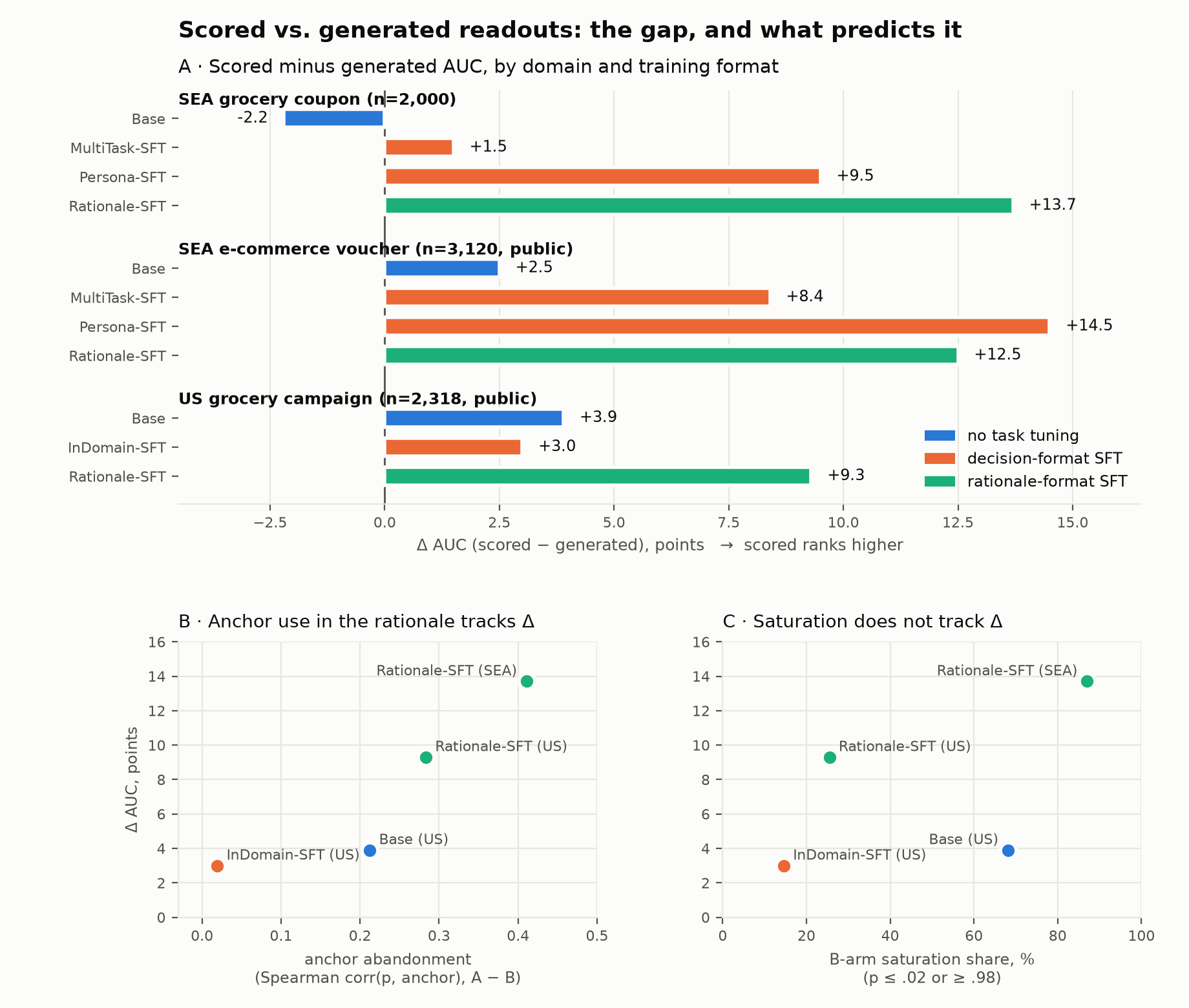}
\caption{(A) $\Delta$AUC (scored $-$ generated) across 11 open-weights cells in three domains,
grouped by domain and colored by training format. (B) $\Delta$ against anchor abandonment:
the drop in rank correlation between the prediction and the dominant prompt feature when a
rationale intervenes. (C) $\Delta$ against probability saturation, which does not order the
cells --- the untuned base saturates most while showing one of the smaller gaps.}
\label{fig:mech}
\end{figure}

Table~\ref{tab:gap} reports the main measurement; Figure~\ref{fig:mech}A visualizes it.

\textbf{Direction and consistency.} The scored readout ranks better in 12 of 13 cells, a
directional consistency a two-sided sign test rejects as chance under the null of no systematic
preference (13 independent cells, $\ge$12 agreeing in direction, $p \approx .003$). The single
exception is the untuned \textbf{Base} on the SEA coupon domain ($-2.2$), which is consistent
with the dose pattern below: a model with no task-specific decision supervision has the least to
lose from routing evidence through text. On the public US grocery domain, where we computed
paired intervals, all three intervals exclude zero.

\textbf{Supervision dose and format mismatch.} The gap is largest for \textbf{Rationale-SFT}
($+9.3$ to $+13.7$), the checkpoint explicitly trained to produce a rationale before deciding.
This is initially counter-intuitive --- the generated readout matches its training format ---
and we return to it in Section~\ref{sec:mechanism}: what rationale-format supervision teaches is a
\emph{confident verdict in a fixed schema}, which transfers poorly when the serving prompt
differs from the trained schema. \textbf{InDomain-SFT}, supervised decision-first on the target
domain's own training split, shows the smallest gap among tuned open-weights models ($+3.0$)
while also achieving the best scored AUC of our checkpoints on that domain ($.808$), indicating
that format-matched in-domain supervision both improves ranking and reduces readout
sensitivity.

\textbf{Order, isolated.} A matched ablation on the voucher domain varies \emph{only} the
position of the decision relative to the rationale: decision-first reaches $.736$ versus $.649$
reasoning-first, i.e.\ $-8.7$ points attributable to ordering alone. On the full voucher test
set the strongest supervised checkpoint moves $.841 \to .775$ when prose precedes the verdict.

\textbf{A degenerate case recovered by readout.} On assortment change, the generated arm is
single-class: all $380$ rows receive the same predicted category, with stated
try-probabilities of $.34$--$.43$ against a $5.3\%$ base rate, making AUC undefined or chance.
Rescoring the \emph{same weights} through the scored readout yields leaves-AUC $.766$. The
information was present in the model; the generated readout did not surface it.

\textbf{The frontier reference.} The API model's number-first cell ($.824$) is the strongest
single result on the public US grocery instrument, exceeding our in-domain checkpoint's scored
readout ($.808$). It nonetheless shows the same directional gap ($+1.4$, interval excluding
zero). We note a practical asymmetry rather than a deficiency: this model exposes neither token
logprobs nor temperature control, so the scored readout --- the better-ranking option for every
open-weights checkpoint we measure --- cannot be constructed against it. Teams building on
reasoning APIs should be aware that readout choice may not be theirs to make.

\subsection{What accompanies the gap}
\label{sec:mechanism}

We dumped every generated rationale together with its conditioned score
($n{=}2{,}000$ SEA coupon; $n{=}2{,}318 \times 3$ US grocery) and measured three quantities per
cell. We report these as \emph{correlates}, established observationally across cells rather
than by intervention; the ordering ablation in Section~\ref{sec:gap} is our one causal manipulation.

\textbf{Verdict saturation is a symptom, not a predictor.} A rationale commits the model:
after 2--4 sentences the decision token concentrates. For Rationale-SFT on the SEA coupon
domain, $87\%$ of conditioned probabilities fall below $.02$ or above $.98$ ($61\%$ and $26\%$
respectively), against a smooth distribution under the scored readout, and the rationale adds
no complementary signal (residual AUC of the generated score given the scored one: $.527$).
But saturation does not order the cells: on US grocery the untuned Base saturates most
($68\%$ of rows, median $p = .010$) while showing one of the smaller gaps ($+3.9$;
Figure~\ref{fig:mech}C).

\textbf{Anchor abandonment tracks the gap.} Each instrument contains a dominant tabular
predictor --- the customer's historical redemption rate, or on US grocery the campaign track
record. Under the scored readout, predictions track it (Spearman $.716$ SEA coupon;
$.739$/$.790$ US grocery). Under the generated readout, rationale-format checkpoints largely
stop using it: correlation falls to $.305$ and $.456$ respectively, and only $17\%$ (SEA) and
$0.2\%$ (US) of rationales reference it explicitly. On the voucher domain the rationale
similarly de-emphasizes the strongest feature (voucher-usage rate, Spearman $.52 \to .19$) in
favor of near-null activity features. Qualitatively, rationales tend to supply a
\emph{normatively plausible} account of the decision rather than the \emph{empirical
regularity} the weights encode --- a shift that is invisible in the text's fluency but visible
in the ranking metric.

\textbf{Template convergence tracks the gap.} Rationale-format checkpoints converge on stock
formulations: the top three opening phrasings cover $19\%$ of $2{,}000$ SEA generations and
$33\%$ of $2{,}318$ US generations, largely in a price-sensitivity register applied regardless
of persona. Near-duplicate rationales among negatives produce tie blocks that reduce ranking
resolution directly.

\textbf{Where the gap is small, the rationale stays anchored.} InDomain-SFT's rank correlation
with the track record barely moves under generation ($.739 \to .720$), its most common openings
explicitly cite that feature, and its template share is $10\%$ --- alongside the smallest gap
among tuned models. Across the cells with per-row data, anchor abandonment and template share
order the gaps correctly (Figure~\ref{fig:mech}B) while saturation does not.

\textbf{Alternative explanations tested.} (i) \emph{Register or tag mismatch}: inserting empty
or neutral \texttt{<think>} blocks leaves scored-level AUC intact ($.745$/$.744$ vs.\ $.748$ on
the voucher domain), so the thinking format itself is not the cause --- the rationale content
is. (ii) \emph{Sampling variance}: self-consistency over three samples recovers only $28\%$ of
the gap. (iii) \emph{Frontier reasoning quality}: the API model's strong generated score on the
SEA coupon domain ($.712$) depends substantially on the anchor feature --- ablating it drops
the model to $.593$ --- and on an anchor-free voucher split the same model scores $.577$, in
the range of our tuned checkpoints under the same readout.

\section{Discussion}
\label{sec:discussion}

We first situate the size of this effect in practical terms (Section~\ref{sec:significance}),
then examine a third readout that partially closes the gap and where it does not
(Section~\ref{sec:repair}), and close with the limitations of our evidence
(Section~\ref{sec:limits}).

\subsection{Why this finding matters in practice}
\label{sec:significance}

\textbf{A costless correction at serving time.} The single most actionable fact in this paper is that
the gap can be closed without retraining, collecting new data, or changing the checkpoint at
all: it is bought back by reading a different token off weights a team already has in
production. On our instruments that switch is worth $3$--$14$ AUC points, a reordering large
enough to materially change who receives a coupon or how a budget is allocated, and it costs
nothing beyond a change to the inference code. The clearest illustration is the
assortment-change domain, where the generated arm looked like a failed deployment --- a single,
uninformative class predicted for every customer --- and the same weights, read differently,
reached leaves-AUC $.766$. A team debugging what looks like a broken or undertrained model may
in fact be looking at a healthy model read through the wrong readout.

\textbf{A retrofittable audit, not just a design recommendation.} Because the two diagnostics
we propose (Section~\ref{sec:protocol}) --- anchor rank correlation and top-3 template share ---
are computed from a rationale dump a team already has, they let an operator estimate how much an
\emph{existing} deployment is paying without running a new experiment, a controlled study, or
access to the original training pipeline. This is what makes the finding useful beyond our own
instruments: it travels to any team that already has a rationale-producing behavioral simulator
in production, regardless of whether they can reproduce our exact checkpoints, and it turns a
question that would otherwise require a bespoke evaluation --- ``is our rationale-based ranking
good enough?'' --- into a five-minute audit. And because the direction of the effect is
consistent across four domains, two continents, proprietary and public data, and open-weights
and frontier-API models alike, a team does not need to take our word for it on faith; the same
two diagnostics let them check whether the pattern holds on their own deployment before they
act on it.

\subsection{A third readout, and where it applies}
\label{sec:repair}

\begin{table}[t]
\centering
\caption{Elicitation isolation on a single checkpoint (MultiTask-SFT, SEA coupon,
$n{=}2{,}000$, base rate $.199$). Requesting a number before any verdict recovers calibration
while ranking within noise of the scored readout.}
\label{tab:isolation}
\small
\begin{tabular}{lccc}
\toprule
Readout & AUC & Brier & mean $P$ \\
\midrule
\armA{} (logprob over YES/NO) & \textbf{.697} & .469 & .759 \\
\armB{} (rationale $\to$ verdict) & .682 & .219 & .213 \\
\armC{} (verbal ``PROB: 0--100'') & .689 & \textbf{.149} & .222 \\
\midrule
\texttt{gpt-5.5}, same \armC{} format & .712 & .143 & .240 \\
\bottomrule
\end{tabular}
\end{table}

Table~\ref{tab:isolation} shows that the choice is not binary between scoring and generating.
Asking for a probability \emph{before} any verdict removes the saturation failure entirely
--- Brier $.469 \to .149$, matching the frontier model's $.143$ --- while ranking within noise
of the scored readout, with no post-hoc calibration layer. The model performs
anchor-and-adjust in a short verbal reply. On US grocery, this readout preserves the anchor
coupling that the rationale readout loses (Spearman $.777$ vs.\ $.456$ for Rationale-SFT).

Two scope conditions bound its use.

\textbf{Rates seen in training.} Asked for per-event aggregate quantities never present in
training --- the share of customers a delisting will cost --- our tuned checkpoints return
essentially flat values (event-level spread $\approx .001$), while the frontier model ranks
events well (Pearson $.894$) with a constant $+17$-point optimism that can be debiased.
Verbalized numbers inherit the training distribution of the register they were trained in;
they do not confer general numeracy.

\textbf{Already-calibrated scored heads.} For InDomain-SFT, whose scored readout is already
well calibrated (Brier $.104$), the verbal-probability readout \emph{costs} $4.6$ AUC points
and improves nothing. This readout is a remedy for miscalibrated scored heads, not a default
upgrade.

\textbf{Training choices that narrow the gap.} Our results support three, for teams who want
rationales in the product:
(i) \emph{in-domain, decision-format supervision} both raises scored AUC and reduces readout
sensitivity (InDomain-SFT: best scored AUC on its domain, smallest gap among tuned models);
(ii) \emph{faithfulness filtering during rationale synthesis} matters --- in our
label-conditioned teacher pipeline, roughly half of draft rationales for negative labels
asserted a disqualifying condition that was factually false for that customer, a
motivated-reasoning artifact removed by requiring every cited fact to appear in the prompt;
(iii) \emph{expect rationales to buy calibration and interpretability rather than ranking}: a
rationale-supervised checkpoint reached $.620$ AUC with Brier $.173$ against $.664$ for its
non-rationale sibling on a matched held-out test, and the decision token inside a trained
rationale is saturated by construction, so graded ranking should still be read from a scored
probe. We regard improving generated-mode ranking --- through faithfulness objectives,
evidence-grounded rewards, or decoupled calibration \citep{dcpo2026} --- as the natural next
step rather than a closed question.

\subsection{Limitations}
\label{sec:limits}

All four domains are retail behavioral prediction with binary outcomes; we do not claim the
result generalizes to other task families, and ranking accuracy is only one of several
properties a deployed simulator is chosen for. Two domains use proprietary data (protocols and
code released; the two public domains carry the reproducible headline numbers). The mechanism
section reports correlates over 11 cells with per-row data on four --- suggestive, not causal,
apart from the ordering ablation. The API cells cannot use the scored readout and run at
provider-default sampling settings, so they compare elicitation order under hidden reasoning
rather than the same three readouts. Historical cells predate our pre-registered replication
and carry their original single-run protocols; a serving-stack change moved absolute AUCs by
$0.5$--$2.4$ points while preserving all orderings, which bounds how precisely cross-run
absolute values should be read. Finally, where tabular baselines are fittable they outperform
all language-model arms on these tasks; the case for a behavioral language model rests on
cold-start coverage, non-tabular signal and the natural-language interface --- including the
generated rationales whose ranking cost we quantify here.

\section{Conclusion}
\label{sec:conclusion}

Behavioral simulators are appealing precisely because a single checkpoint appears to deliver
both a prediction and its explanation, and practitioners have had little reason to doubt that
the two are free of each other. We show that they are not: across 13 model-domain cells
spanning four retail prediction tasks, reading a decision probability from the model ranks real
customer outcomes more accurately than asking it to reason first and then answer, a directional
pattern too consistent to be chance (12 of 13 cells, two-sided sign test $p\approx.003$) and too
large to ignore in a targeting decision (up to $14.5$ AUC points). The gap is not a fixed tax on
generation; it is a predictable function of how a checkpoint was trained, growing with
task-specific supervision and with mismatch between the training and serving formats, and it is
traceable at the level of individual rationales to two concrete behaviors --- abandoning the
dominant predictive feature and converging on stock phrasing --- rather than to the verdict
saturation that is most visible to the eye. A third readout, verbalized probability elicited
before any verdict, shows that the choice is not binary: it recovers calibration without paying
the full ranking cost, though only within the rates a checkpoint has already seen in training
and only when its scored head was miscalibrated to begin with.

Taken together, these results reframe the design question for teams building behavioral
simulators. The rationale a model produces is not a free byproduct of a prediction; it is a
different measurement, drawn through a narrower channel, and it should be evaluated and priced
as such. Nothing here argues against generating rationales --- they remain the interface that
makes a behavioral simulator auditable and usable in the first place --- only against trusting
them, unaudited, for the ranking decision itself. What makes this actionable rather than merely
diagnostic is that the fix requires no new data or retraining: the same checkpoint a team
already runs in production yields $3$--$14$ additional AUC points simply by reading a different
token off it, and the same two diagnostics we used to find the gap (Section~\ref{sec:significance})
let any operator check, in minutes, how much their own deployment is paying before deciding
whether to act. The remaining question, which we leave open, is whether the ranking cost of
generation can be trained away directly, through faithfulness objectives or evidence-grounded
rewards, rather than routed around at serving time as we recommend below.

\section{Deployment guidance}
\label{sec:protocol}

Our results support five practical recommendations for teams deploying a behavioral language
model. First, \textbf{source ranking from the scored readout} for targeting, ordering and
budget allocation; on our instruments this is worth $3$--$14$ AUC points over generating first.
Second, \textbf{keep the rationale, but condition it on the decision}: generating an
explanation \emph{after} the scored decision retains the interpretability surface at no
measured ranking cost, since the text cannot feed back into the score. Third, \textbf{use the
verbal-probability readout for rate forecasts} when the scored head is miscalibrated and an
anchor-style feature is present in the prompt, validating on the trained register before
extending it to novel aggregates. Fourth, \textbf{prefer in-domain decision-format supervision}
where the target task's own training split is available, since it improved both ranking and
readout robustness in our measurements. Fifth, \textbf{diagnose an existing deployment} by
dumping ${\sim}2$k rationales and measuring two quantities: the rank correlation between the
prediction and the dominant feature (scored vs.\ generated), and the top-3 template share; in
our cells these two diagnostics order the gap correctly, while saturation does not.

\subsubsection*{Reproducibility statement}
The US grocery domain is fully public: dataset (public mirror; one documented header rename
reproduces our instrument exactly --- $2{,}318$ test rows, $14.0\%$ positive), checkpoints,
evaluation and analysis code, and per-row dumps. The voucher domain uses the public DMBGN
benchmark. Prompts for all readouts, the pinned protocol for each historical cell, and the
bootstrap scripts are in the supplement. Proprietary-domain protocols are released without
row-level data.

\subsubsection*{Ethics statement}
All customer data is pseudonymized loyalty-program data or public benchmark data; no row-level
proprietary data leaves the training environment, and the retailer is not named. Behavioral
simulators can support manipulative targeting; we note that a readout which preserves
calibration makes over-confident targeting more visible to the operator, and that the
diagnostics in Section~\ref{sec:protocol} are usable by auditors as well as builders.

\bibliography{references}
\bibliographystyle{iclr2026_conference}

\appendix
\section{Prompts and readout templates}
\label{app:prompts}
Verbatim system and user templates per domain and readout, including the \texttt{<think>} cue,
the YES/NO cue and the probability cue; the substitution mapping one prompt to another; parser
regular expressions and failure counts.

\section{Training details}
\label{app:training}
Per-checkpoint LoRA rank and target modules, optimizer and schedule, sequence packing, corpus
composition tables (CPT source mix and token counts; SFT lane counts by task family), the
replay fraction ablation, and the retrieval-in-CPT negative result.

\section{Per-cell protocol ledger}
\label{app:ledger}
Decoding mode, shard and seed counts, serving stack (inference engine version, attention
backend) per cell, and the cross-stack drift measurement.

\section{Instrument construction}
\label{app:instruments}
Detection rules for the assortment-change natural experiments, the defect ledger including the
withdrawn introduction family, the US grocery time-split and leak audit, and the
customer-disjoint construction of the SEA coupon instrument.

\end{document}